\documentclass[sigconf,nonacm]{acmart}
\AtBeginDocument{%
  }

\usepackage{multirow}
\usepackage{enumitem}
\graphicspath{{figures/}}

\begin{document}
\newcommand{\name}{Remember-R1}


\title{\name: Mitigating Long-Context Visual Forgetting through Reinforcement Learning}


\author{Jianmin Chen}
\authornote{The first two authors contributed equally.}
\affiliation{%
  \institution{Northwestern Polytechnical University}
  \city{Xi'an}
  \country{China}
}

\author{Jiaqi Tang}
\authornotemark[1]
\affiliation{%
  \institution{Hong Kong University of Science and Technology}
  \city{Hong Kong}
  \country{China}
}

\author{Wei Wei}
\authornote{Corresponding author.}
\affiliation{%
  \institution{Northwestern Polytechnical University}
  \city{Xi'an}
  \country{China}
}

\author{Xiaogang Xu}
\affiliation{%
  \institution{Zhejiang University}
  \city{Hangzhou}
  \country{China}
}

\author{Jiafei Wu}
\affiliation{%
  \institution{Zhejiang University}
  \city{Hangzhou}
  \country{China}
}

\author{Zhe Liu}
\affiliation{%
  \institution{Zhejiang University}
  \city{Hangzhou}
  \country{China}
}

\author{Qianzhou Wang}
\affiliation{%
  \institution{Northwestern Polytechnical University}
  \city{Xi'an}
  \country{China}
}

\author{Yingying Yan}
\affiliation{%
  \institution{Northwestern Polytechnical University}
  \city{Xi'an}
  \country{China}
}

\author{Botong Geng}
\affiliation{%
  \institution{Northwestern Polytechnical University}
  \city{Xi'an}
  \country{China}
}

\author{Yuyang Xia}
\affiliation{%
  \institution{Northwestern Polytechnical University}
  \city{Xi'an}
  \country{China}
}

\author{Lei Zhang}
\affiliation{%
  \institution{Northwestern Polytechnical University}
  \city{Xi'an}
  \country{China}
}

\author{Qifeng Chen}
\affiliation{%
  \institution{Hong Kong University of Science and Technology}
  \city{Hong Kong}
  \country{China}
}


\renewcommand{\shortauthors}{Chen et al.}
\settopmatter{authorsperrow=3}

\begin{abstract}
Multimodal large language models (MLLMs) increasingly rely on long chain-of-thought reasoning for complex tasks. However, as reasoning sequences lengthen, models may gradually rely less on visual evidence and more on accumulated textual context, leading to visual forgetting. Existing approaches do not directly constrain how visual evidence is used and maintained along the original reasoning trajectory, leaving long-context visual forgetting insufficiently addressed. To address this issue, we propose \name, a reinforcement learning framework that mitigates long-context visual forgetting by applying process-level supervision directly on the original reasoning trajectory. Specifically, \name\ introduces rewards that encourage broader coverage of matched visual keywords, stronger persistence of visual dependence in later reasoning steps, and greater focus on question-relevant image regions. Experiments across multiple model scales and diverse multimodal benchmarks demonstrate that \name\ consistently improves reasoning performance. Additional analyses further show that it slows the decline of visual attention during generation, supporting its effectiveness in mitigating long-context visual forgetting. The code is available: https://github.com/Ch921-cell/Remember-R1.
\end{abstract}



\maketitle

\begin{figure}[t]
    \centering
    \vspace{1.5em}
    \includegraphics[width=1\linewidth]{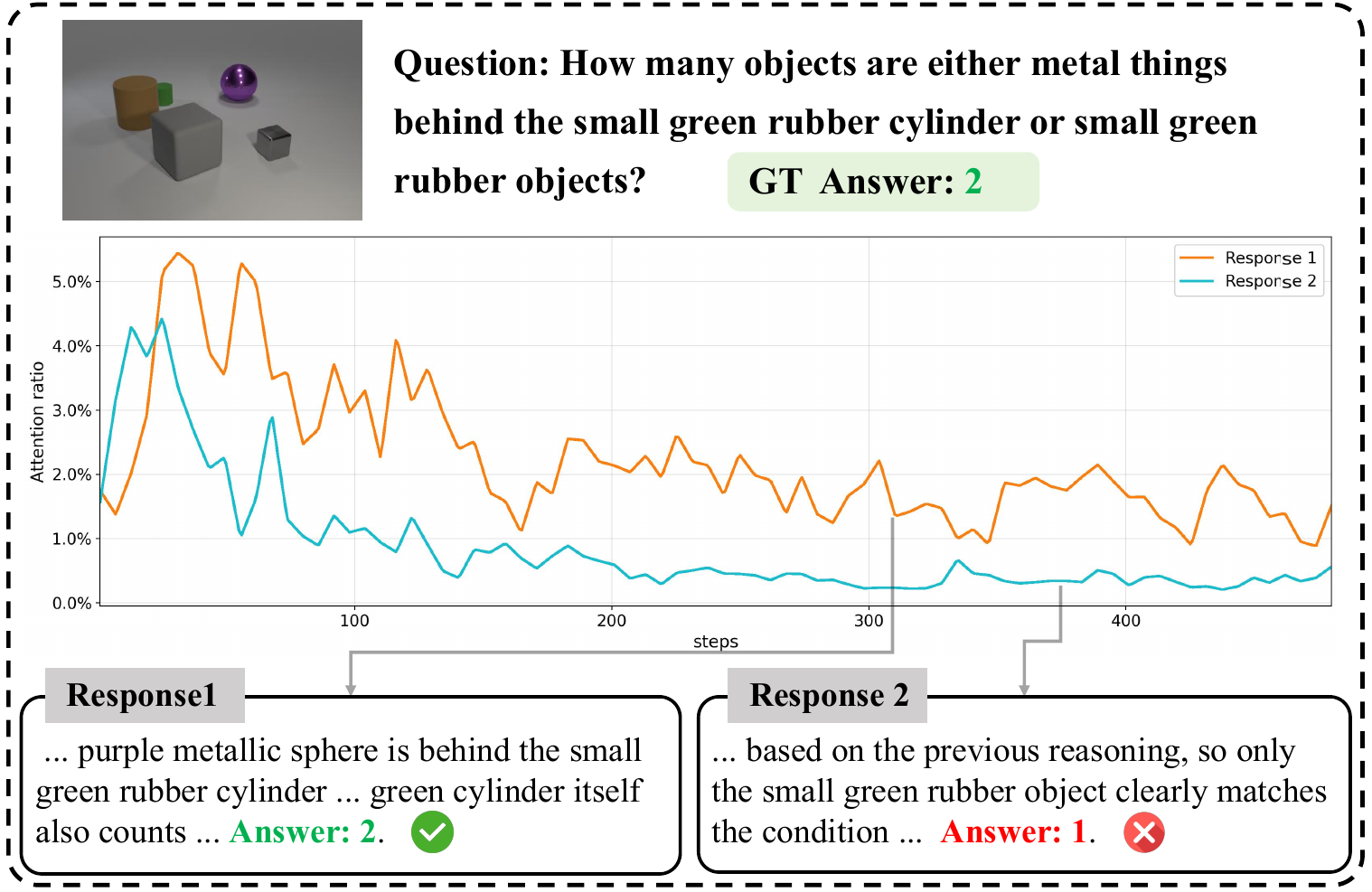}
    \caption{Example of visual forgetting in long-context multimodal chain-of-thought reasoning. For the same image-question pair, we compare two reasoning responses by tracking the ratio of attention assigned to visual tokens across the generation steps.}
    \vspace{-2em}
    \label{fig:introduction}
\end{figure}

\begin{figure}[t]
    \centering
    \includegraphics[width=1\linewidth]{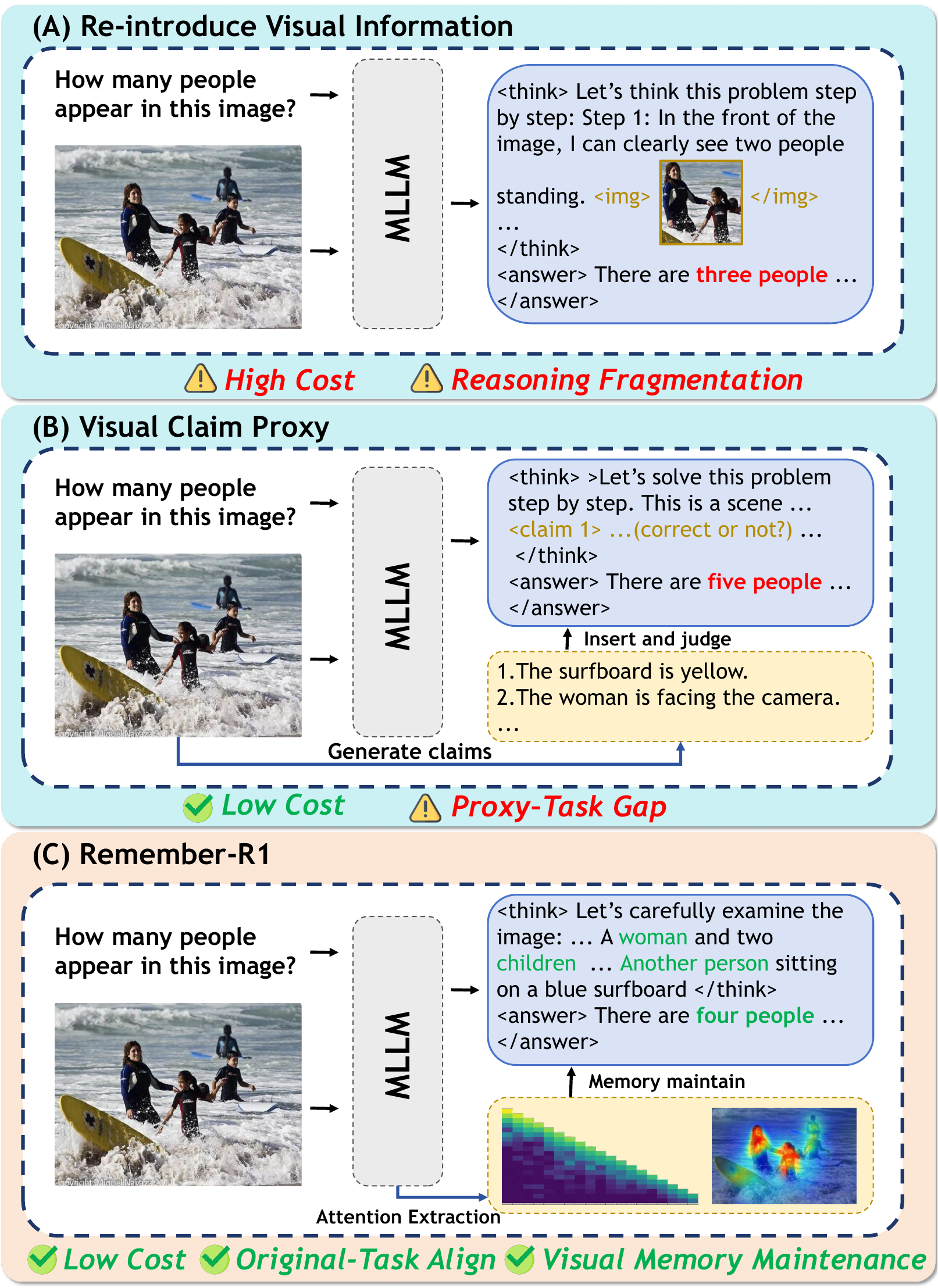}
    \caption{Comparison of three approaches to mitigating visual forgetting. (A) Visual re-introduction increases inference cost. (B) Visual-claim proxies introduce a proxy-task gap. (C) \name\ applies process-level supervision to the original reasoning trajectory without changing inference.}
    \vspace{-1.5em}
    \label{fig:Motivation}
\end{figure}
\begingroup
\setlength{\emergencystretch}{0.75em}
\section{Introduction}

MLLMs are rapidly shifting from perception-oriented tasks, such as image captioning and visual question answering, to tasks requiring multi-step reasoning over visual inputs~\citep{wei2022chain, zhang2023multimodal, tang2026robust, tang2026robustu1, ma-etal-2026-response, NEURIPS2024_fca83589, tang2026intelligent}. Recent work on multimodal chain-of-thought reasoning shows that longer reasoning traces can improve performance on some challenging benchmarks, including mathematical reasoning, logical reasoning, and document understanding~\citep{liu2024deepseek,jaech2024openai}. 

However, longer reasoning traces also expose a key weakness of current MLLMs: \textbf{visual forgetting during reasoning}. Recent studies~\citep{sun2025mitigating} show that, as reasoning proceeds, the model attends less to the image; later tokens are driven more by previously generated text than by visual evidence. As a result, the reasoning chain may drift away from the visual facts. This issue is particularly consequential in multi-step mathematical, logical, and spatial reasoning, which often requires longer chains that must remain tightly aligned with visual facts throughout the process.

Prior work has used attention weights as a useful diagnostic signal of model focus~\citep{kang2025your, zhang2025mllms}. In the same spirit, we track attention to visual information to examine how the model’s use of the image evolves over the reasoning process. Figure~\ref{fig:introduction} compares two reasoning responses to the same image-question pair. The correct response maintains a relatively higher visual attention ratio in later steps and continues to reference visual facts, whereas the incorrect response shows a sharper late-stage decline in visual attention and increasingly relies on previously generated text rather than on the image. This example suggests a phenomenon of visual forgetting, revealing the model’s limitations in maintaining consistent attention to visual information throughout the reasoning process.

Existing approaches to mitigating visual forgetting fall into two broad categories. One line of work, shown in Figure~\ref{fig:Motivation}(A), addresses forgetting through visual re-introduction, where the original image or some regions of it are introduced back into the context~\citep{yang2026look,sun2025mitigating}. This can improve long-form reasoning, but repeated visual processing increases inference cost in both computation and memory, and may also lead to reasoning fragmentation.

A second line of work, shown in Figure~\ref{fig:Motivation}(B), improves visual grounding by inserting visual claim proxies into the reasoning process~\citep{tian2025more}. However, the resulting supervision is applied to these added proxy interactions rather than directly to the original target rollout, and therefore does not directly constrain how visual evidence is maintained along the untouched reasoning trajectory.

What remains missing is direct supervision on the original reasoning trajectory, where visual forgetting emerges. As the model proceeds through a long reasoning chain, its dependence on visual evidence may gradually weaken, leading to visual forgetting in later reasoning steps. Supervising the use of visual evidence on that same trajectory therefore provides a more direct way to encourage sustained grounding throughout response generation.

Motivated by this, we propose \textbf{\name}, an RL framework that places process-level supervision on the target rollout itself without modifying the inference procedure. \name\ uses three complementary rewards to encourage broader coverage of matched annotated visual keywords, maintain visual dependence into later reasoning steps, and focus that dependence on question-relevant image regions. In this way, the training signal acts directly on the trajectory where forgetting develops, rather than through auxiliary inserted interactions.

\begin{figure*}[t]
\centering
\includegraphics[width=0.32\linewidth]{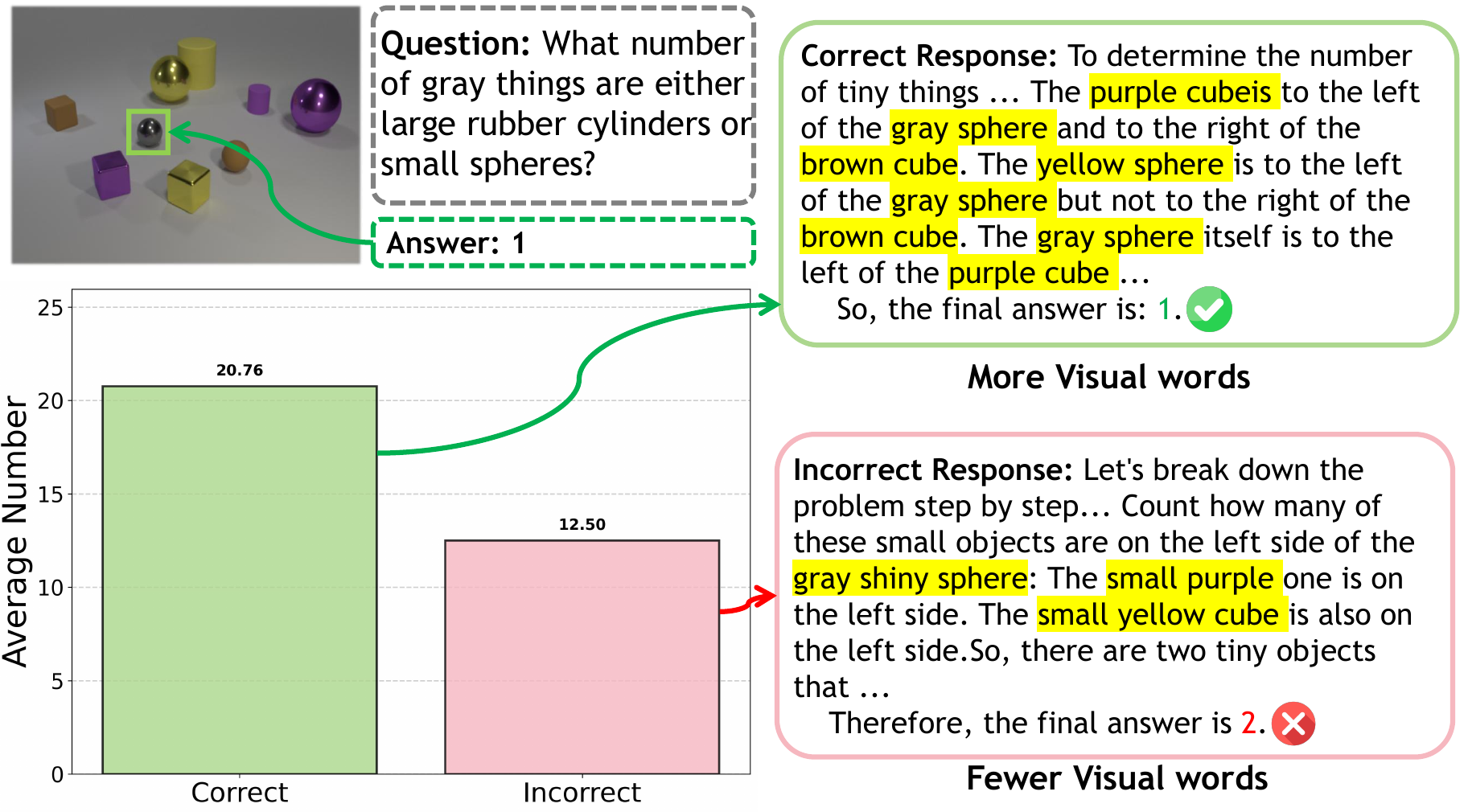}
\hfill
\includegraphics[width=0.32\linewidth]{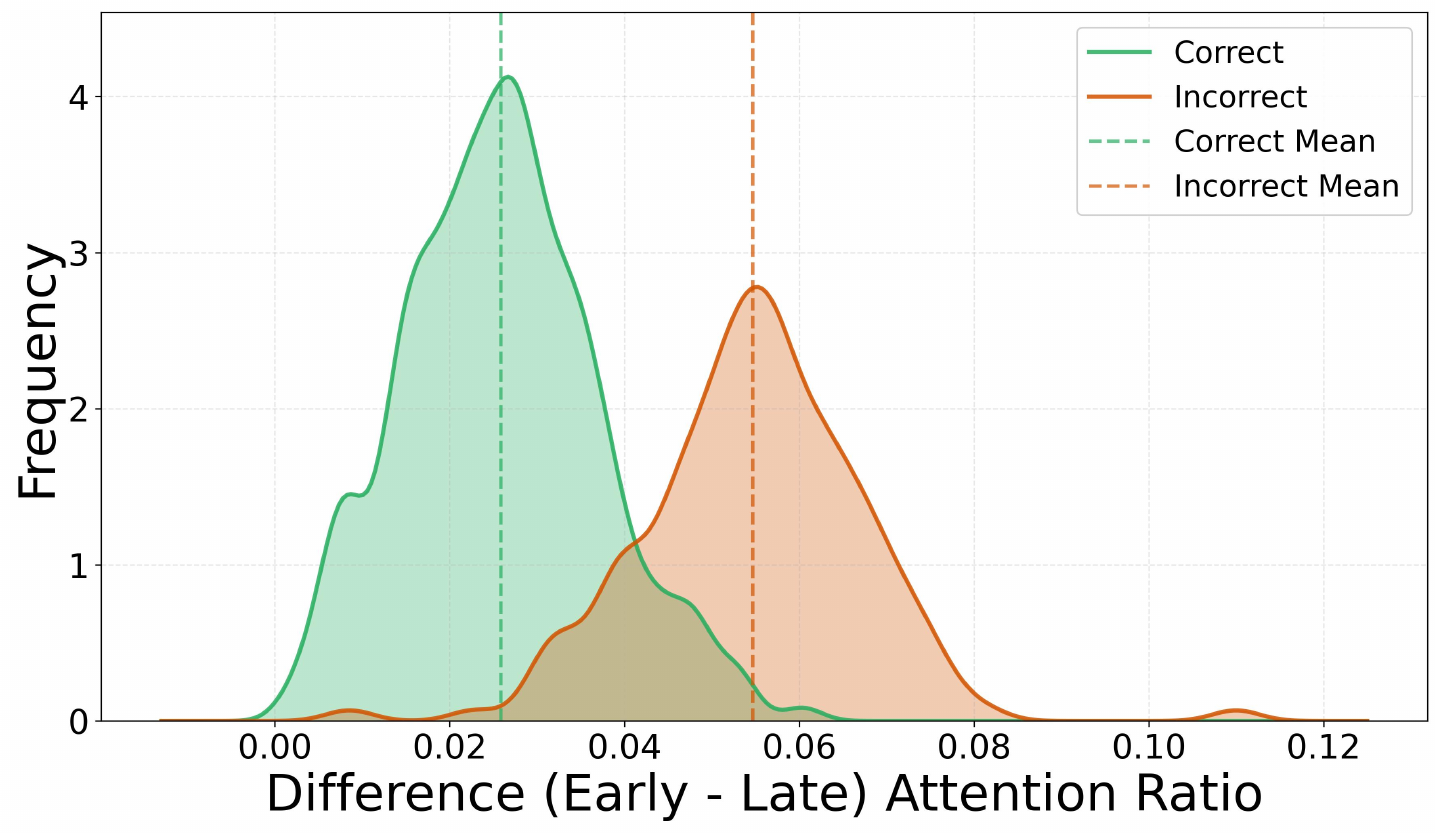}
\hfill
\includegraphics[width=0.32\linewidth]{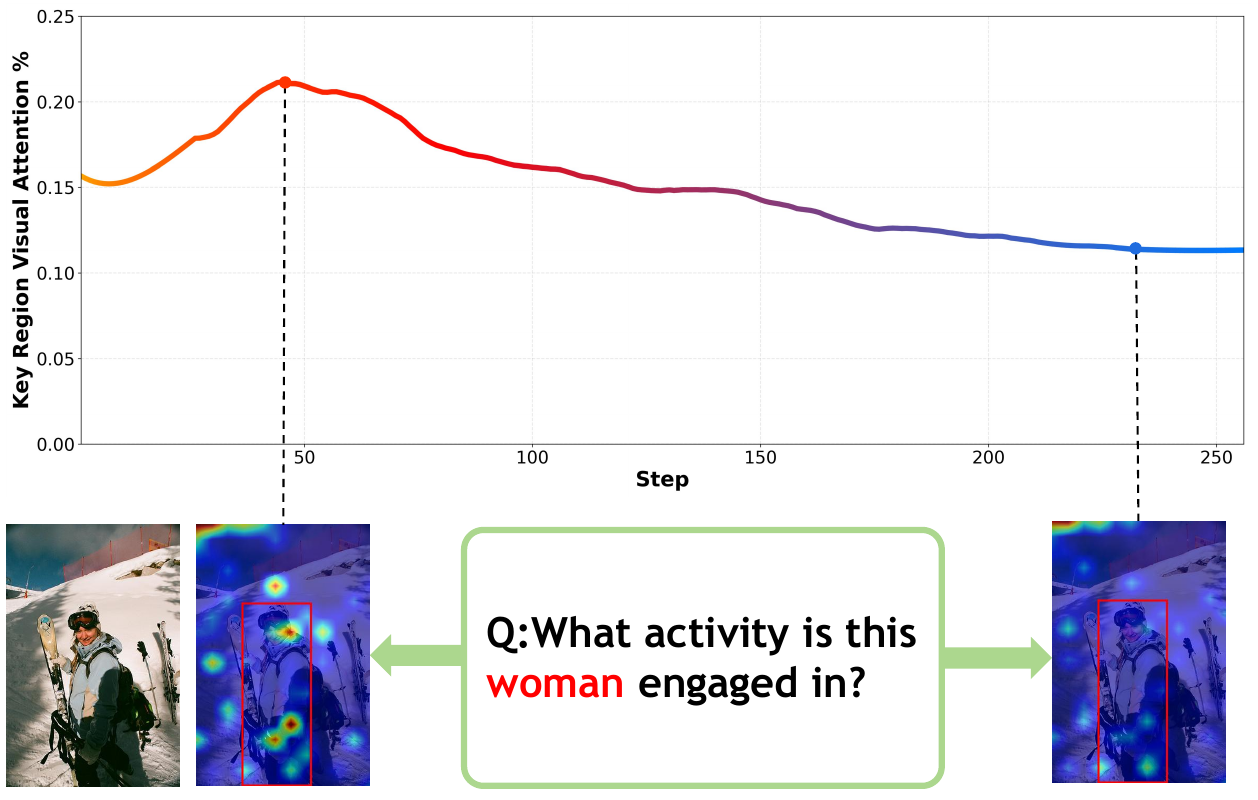}
\par
\makebox[0.32\linewidth][c]{\small (a) Visual Keyword Usage}
\hfill
\makebox[0.32\linewidth][c]{\small (b) Visual Attention Decay}
\hfill
\makebox[0.32\linewidth][c]{\small (c) Key-Region Attention}
\caption{Benchmark-level evidence for the three-aspect decomposition underlying our reward design. (a) Correct responses tend to contain more matched annotated visual keywords. (b) Correct responses exhibit smaller early-to-late declines in visual attention. (c) Correct responses maintain stronger attention on question-relevant regions over reasoning steps.}
\label{fig:motivation_study}
\end{figure*}

We evaluate \name\ in three settings: mathematical and logical reasoning, general multimodal capability, and visual perception. Experimental results show that \name\ consistently improves performance across these evaluations. Additional analyses show that it slows the decline of visual attention during generation, particularly in the middle and later stages. These results show that by directly supervising visual evidence use on the target reasoning trajectory, \name\ helps the model better maintain reliance on visual evidence as generation unfolds, thereby mitigating long-context visual forgetting.

Our main contributions are summarized as follows:
\begin{itemize}[leftmargin=*,topsep=0.25em,itemsep=0.2em,parsep=0pt,partopsep=0pt]
    \item We study long-context visual forgetting from the perspective of directly supervising the original rollout and propose a process-level supervision framework that does not modify the inference procedure.

    \item We design three process-level rewards that supervise complementary aspects of visual evidence use during reasoning: coverage of matched annotated visual keywords, persistence of visual attention over later reasoning steps, and focus on question-relevant key regions.
    
    \item We validate \name\ across two model scales and seven benchmarks; ablation and attention analyses show improvements in both overall performance and visual-evidence preservation.
\end{itemize}
\endgroup

\vspace{-1.5em}
\section{Related Work}


\textbf{Visual Forgetting.}
Recent work has begun to examine how MLLMs maintain reliance on visual evidence during long reasoning chains~\citep{sun2025mitigating,zheng2025deepeyes,wu2026iqat1toolbasedvisualevidence}. Existing methods mainly follow two directions. The first uses
visual re-introduction, where the image or visual features are
reintroduced during inference~\citep{yang2026look}. This can help recover visual evidence during generation, but repeated visual processing
increases inference cost and may also interrupt the continuity of the reasoning process. The second direction improves visual grounding through proxy interactions during training~\citep{tian2025more}. These methods provide additional supervision, but that supervision is not defined directly on the original reasoning trajectory of the target problem. Our method differs from both directions: it leaves inference unchanged and applies reward signals directly to the target reasoning trajectory itself.

\textbf{Reinforcement Learning for MLLMs.}
Reinforcement learning has been widely used to improve reasoning and response quality in both large language models and multimodal large language models. Earlier methods mainly relied on PPO-style optimization, while later approaches such as DPO and GRPO simplified training or improved reasoning performance~\citep{schulman2017proximal, rafailov2023direct, shao2024deepseekmath, liu2024deepseek, tang-etal-2026-lpo}. In multimodal models, prior RL methods have mostly used rewards defined on final answers or overall outputs~\citep{sun2024aligning}. More recent work has begun to explore process-level supervision for multimodal reasoning~\citep{li2026palmr, khalifa2026process}. In our setting, such supervision is applied directly to the original reasoning trajectory to encourage sustained reliance on visual evidence throughout long-form generation.

\section{Methodology}

\textbf{Preliminaries.}
Consider a multimodal large language model (MLLM) parameterized by $\theta$. Let $\mathcal{I}$ denote the visual input and $\mathcal{Q}$ denote the question. Given $(\mathcal{I}, \mathcal{Q})$, the model generates a response $Y=(y_1,y_2,\dots,y_T)$, which includes both the reasoning steps and the final answer, under the standard autoregressive formulation:
\begin{equation}
P_\theta(Y \mid \mathcal{I}, \mathcal{Q})
=
\prod_{t=1}^{T}
P_\theta(y_t \mid \mathcal{I}, \mathcal{Q}, y_{<t}),
\end{equation}
where $y_{<t}$ denotes the previously generated tokens.

\textbf{Quantifying Visual Forgetting.}
To provide a quantitative basis for the process-level rewards introduced below, we use attention to visual tokens as a proxy for the model’s reliance on visual evidence during generation~\citep{kang2025your, zhang2025mllms}. Let \(\Omega(t)\in[0,1]\) denote the aggregate attention mass assigned to visual tokens at generation step \(t\). Although \(\Omega(t)\) does not fully characterize model behavior, it provides an observable signal of how reliance on visual evidence changes over the reasoning trajectory. We capture visual forgetting through the early-to-late change in visual attention,
\[
\Delta_{\mathrm{vis}}
=
\mathbb{E}_{t\in T_{\mathrm{early}}}[\Omega(t)]
-
\mathbb{E}_{t\in T_{\mathrm{late}}}[\Omega(t)],
\]
where a larger value indicates a stronger late-stage decline in visual attention. This view motivates the process-level rewards introduced below.
\begin{figure*}[t]
    \centering
    \includegraphics[width=1.0\linewidth]{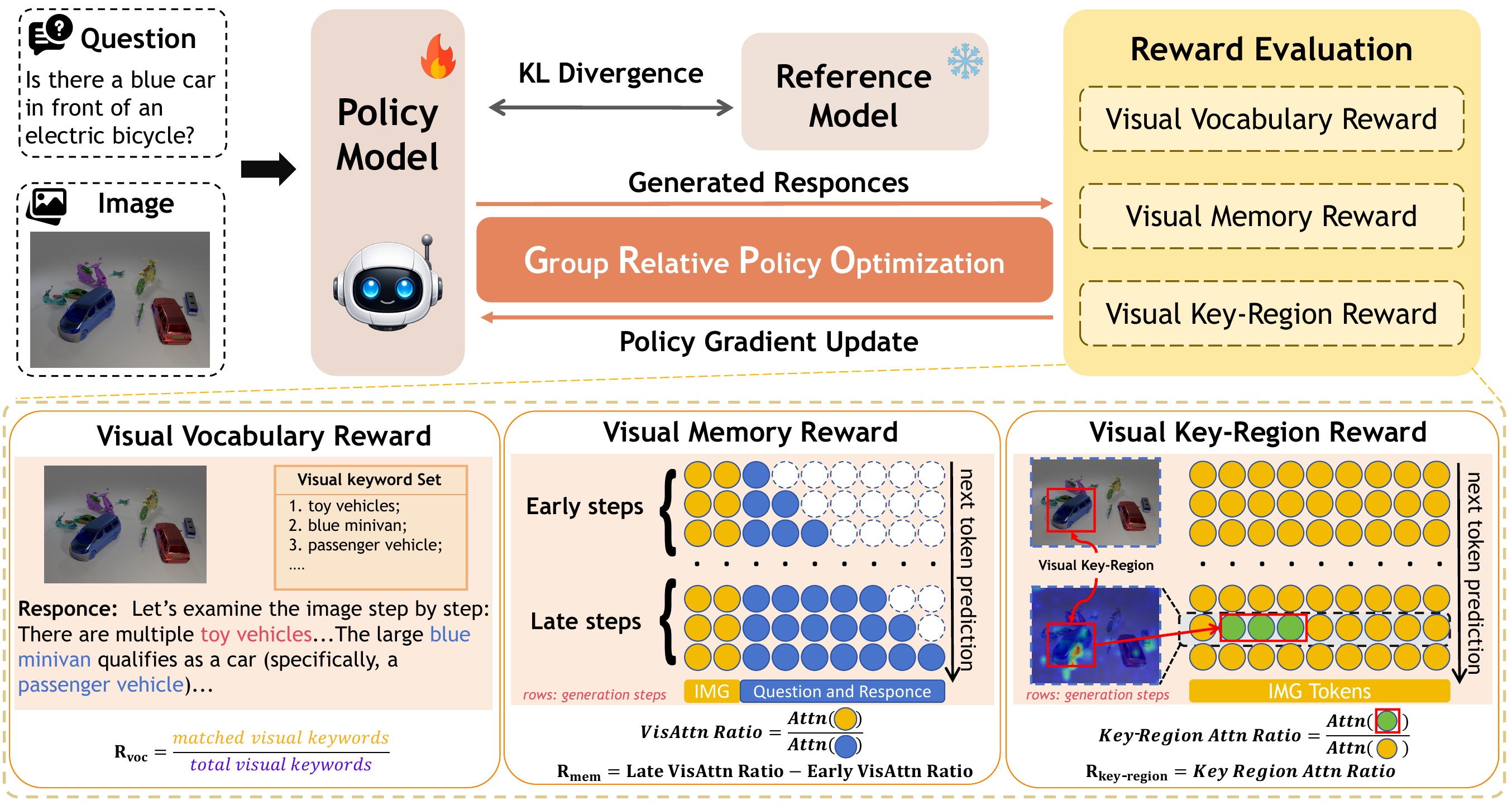}
    \caption{Overview of \name. Three process-level rewards supervise matched annotated visual-keyword coverage, visual-attention persistence, and region relevance during training, while inference remains unchanged.}
    \label{fig:Architecture}
\end{figure*}

\subsection{Visual Evidence Annotation}
\label{subsec:data_construction}

To compute process-level rewards on the original reasoning trajectory, we augment the ViRL39K dataset~\citep{wang2025vl} with structured annotations of visual evidence for each training sample. These annotations provide the supervision needed for the annotation-dependent rewards defined below.

\textbf{Visual Keyword Extraction.}
For each sample $i$, we extract a set of visual keywords from the image $I_i$ using Qwen-VL-Max~\citep{yang2025qwen3}, denoted as $\mathcal{M}$:
\begin{equation}
    \mathcal{K}_i = \{k_{1}, \dots, k_{N}\} = \mathcal{M}(I_i).
    \label{eq:visual_vocab}
\end{equation}
Each keyword is a short lexical unit grounded in visible content, including object categories (e.g., \textit{cube}, \textit{sphere}), attributes (e.g., \textit{blue}, \textit{small}), quantities, and short spatial or compositional phrases when necessary (e.g., \textit{leftmost bar}, \textit{yellow object on the right}). We keep the keywords short so that they can serve as compact supervision for explicit mention of visual evidence during reasoning, rather than as full-sentence descriptions.

\textbf{Key-Region Box Annotation.}
For the same sample $i$, we further identify image regions that provide the visual evidence needed to answer the question. Specifically, we query $\mathcal{M}$ with $(I_i, Q_i, A_i)$ and obtain
\begin{equation}
    \mathcal{B}_i = \{b_{1}, \dots, b_{K}\} = \mathcal{M}(I_i, Q_i, A_i),
    \label{eq:key_region}
\end{equation}
where each box $b_k = (x_k, y_k, w_k, h_k)$ specifies the center coordinates, width, and height of one annotated region.

Importantly, $A_i$ is used solely in this one-time annotation pipeline and is never exposed to the policy model as an input or conditioning signal during reinforcement learning or inference. The policy always operates on the standard input pair $(I_i, Q_i)$ and the annotated boxes are used only by the external reward evaluator after a rollout is generated.

\textbf{Manual Verification.}
We manually verify both $\mathcal{K}_i$ and $\mathcal{B}_i$ to improve annotation quality. Human remove redundant or visually unsupported keywords, correct obvious category/attribute mismatches, adjust inaccurate boxes, and discard or regenerate samples whose annotations remain ambiguous after correction. This step is important because annotation quality directly affects the reliability of the vocabulary-based and region-based rewards.

\textbf{Dataset Summary.}
After filtering and manual verification, the final dataset contains 38,657 samples, each annotated with visual keywords $\mathcal K_i$ and key-region boxes $\mathcal B_i$.

\subsection{Reward Design}

A decline in visual attention provides a useful diagnostic signal of visual forgetting, but it does not by itself specify how supervision should be imposed on the reasoning trajectory. To translate this diagnosis into trainable objectives, we decompose sustained reliance on visual evidence during reasoning into three complementary aspects: coverage of matched annotated visual keywords, persistence of visual attention into later reasoning steps, and concentration of visual attention on regions relevant to the question. Based on this decomposition, \name\ uses three corresponding process-level rewards, as shown in Figure~\ref{fig:Architecture}.

Figure~\ref{fig:motivation_study} provides benchmark-level evidence for each aspect of this decomposition. Across benchmark examples, correct responses tend to contain more matched annotated visual keywords, exhibit smaller early-to-late declines in visual attention, and maintain stronger attention on question-relevant regions.

\subsubsection{Visual Vocabulary Reward}

One aspect of mitigating visual forgetting is the explicit mention of annotated visual keywords in the reasoning trace. Figure~\ref{fig:motivation_study}(a) shows that correct responses contain substantially more matched annotated visual keywords than incorrect ones (20.76 vs.\ 12.50 on average), motivating the use of keyword coverage as a supervision signal. Accordingly, we introduce the \emph{Visual Vocabulary Reward} to encourage the model to mention these annotated visual keywords during generation.

\textbf{Keyword matching.}
For each example, let $\mathcal{K}$ denote the precomputed visual keyword set constructed in Eq.~\eqref{eq:visual_vocab}. Given a generated response $Y$, we first lowercase the response and keywords and remove minor formatting differences. We then perform exact phrase matching between the response and the annotated keyword set. A keyword is counted only if it appears as an exact standalone word or phrase, and each keyword contributes at most once. We do not use synonym expansion. We define
\begin{equation}
\mathcal{K}_i^{\mathrm{match}}(Y)
=
\{k \in \mathcal{K}_i : k \text{ is matched in } Y\}.
\end{equation}

\textbf{Formulation.}
The reward is computed as a clipped average of position-weighted matched keywords. For sample $i$, the \textbf{Visual Vocabulary Reward} is defined as
\begin{equation}
r_{\mathrm{voc}}(Y) = \min \left( 1,\; \frac{c}{|\mathcal{K}_i|} \sum_{k \in \mathcal{K}_{i}^{\mathrm{match}}(Y)} \frac{\tau(k)}{T_{\mathrm{eff}}} \right),
\end{equation}
where $\tau(k)$ denotes the token position of the last matched occurrence of keyword $k$ in the response, $T_{\mathrm{eff}}=\max(T,L_{\min})$ is an effective response length used for normalization, and $c$ is a scaling coefficient controlling the reward magnitude before clipping.

This formulation serves two purposes. First, the summation over $\mathcal{K}_{i}^{\mathrm{match}}(Y)$ encourages broader coverage of matched annotated visual keywords rather than only a small subset. Second, weighting each matched keyword by its last occurrence position gives relatively higher reward when matched annotated visual keywords appear in later parts of the response. This weighting is intended to encourage continued reference to annotated visual keywords as reasoning unfolds, rather than concentrating all such references near the beginning of the response.

\subsubsection{Visual Memory Reward}

Explicit mention of matched annotated visual keywords alone is insufficient if the model gradually stops relying on visual information as reasoning proceeds. Figure~\ref{fig:motivation_study}(b) shows that incorrect responses exhibit larger early-to-late declines in attention to visual information, whereas correct responses show smaller declines between the early and late stages of generation. This pattern motivates limiting late-stage attention decay during multimodal reasoning. We therefore define the \emph{Visual Memory Reward} to discourage substantial late-stage decay in attention to visual information.

\textbf{Attention extraction.}
For a generated response $Y$ of length $T$, let $\mathcal{T}_v$ denote the set of visual token indices in the model input. Using attention weights from the final transformer layer, we define the total visual attention at generation step $t$ as
\begin{equation}
\Omega(t) = \frac{1}{H} \sum_{h=1}^{H} \sum_{j \in \mathcal{T}_v} A_t^{h}(j),
\label{eq:visual_attention}
\end{equation}
where $A_t^{h}(j)$ denotes the attention weight from the token generated at step $t$ to visual token $j$ under attention head $h$.

\textbf{Formulation.}
To quantify how well visual attention is maintained from early to late stages of generation, we define a window size
\begin{equation}
w = \max(1,\lfloor \gamma T \rfloor),
\end{equation}
where $\gamma$ is a fixed fraction of the response length. We then compute the average visual attention over the first and last windows:
\begin{equation}
\mu_{\mathrm{start}} = \frac{1}{w}\sum_{t=1}^{w}\Omega(t), \qquad
\mu_{\mathrm{end}} = \frac{1}{w}\sum_{t=T-w+1}^{T}\Omega(t).
\end{equation}

The \textbf{Visual Memory Reward} is defined as
\begin{equation}
r_{\mathrm{mem}}(Y) = 1 + \left(\mu_{\mathrm{end}} - \mu_{\mathrm{start}}\right).
\end{equation}
This reward is larger when late-stage visual attention remains closer to, or exceeds, its initial level and smaller when it declines substantially between the initial and final windows. In this way, it encourages higher late-stage visual attention relative to the initial stage.

\subsubsection{Visual Key-Region Reward}

Maintaining overall reliance on visual tokens is still insufficient if the preserved attention is not focused on image regions that are relevant to the question. Figure~\ref{fig:motivation_study}(c) shows that the proportion of visual attention assigned to annotated key regions may still decrease over generation even when total visual attention remains relatively stable. We therefore define the \emph{Visual Key-Region Reward} to encourage sustained attention to question-relevant regions during reasoning.

\textbf{Box-to-token mapping.}
For a training sample $(I_i, Q_i)$, let $\mathcal{B}_i=\{b_1,\dots,b_K\}$ denote the annotated key-region boxes defined in Eq.~\eqref{eq:key_region}. For patch-based vision encoders, each visual token corresponds to a spatial patch in the input image. We associate each visual token $j \in \mathcal{T}_v$ with the center coordinate of its corresponding patch, denoted by $c(j)$. The token set associated with box $b_k$ is defined as
\begin{equation}
S(b_k) = \{\, j \in \mathcal{T}_v : c(j) \in b_k \,\},
\end{equation}
and the union over all annotated boxes is
\begin{equation}
S_{\mathrm{union}} = \bigcup_{k=1}^{K} S(b_k).
\end{equation}
This yields a deterministic mapping from image-space annotations to visual tokens.

\textbf{Region-attention ratio.}
Using the same final-layer attention weights, we define the attention mass assigned to the annotated key regions at generation step $t$ as
\begin{equation}
\Omega_{B}(t) = \frac{1}{H}\sum_{h=1}^{H}\sum_{j \in S_{\mathrm{union}}} A_t^{h}(j).
\end{equation}
Let $\Omega(t)$ denote the total attention mass assigned to all visual tokens at step $t$, as defined in Eq.~\eqref{eq:visual_attention}. We then compute the region-attention ratio as
\begin{equation}
R_B(t) = \frac{\Omega_B(t)}{\Omega(t)+\epsilon_{\mathrm{den}}},
\end{equation}
where $\epsilon_{\mathrm{den}}$ is a small constant for numerical stability.

\textbf{Formulation.}
The \textbf{Visual Key-Region Reward} is defined as
\begin{equation}
r_{\mathrm{region}}(Y) = \frac{1}{Z}\sum_{t=1}^{T}\left(\frac{t}{T}\right) R_B(t),
\end{equation}
where
\begin{equation}
Z = \sum_{t=1}^{T}\frac{t}{T}.
\end{equation}
This formulation computes a temporally weighted average of the proportion of visual attention assigned to question-relevant regions, with larger weights assigned to later generation steps. It therefore assigns higher reward when the model better preserves question-relevant visual grounding in later stages of reasoning.

\subsection{Reinforcement Learning}

\textbf{Total reward.}
We combine the answer correctness reward with the three visual rewards. For a generated rollout $Y$, the total reward is
\begin{equation}
\mathcal{R}(Y,\mathcal{I},\mathcal{Q}) =
r_{\mathrm{acc}}(Y)
+
r_{\mathrm{voc}}(Y)
+
r_{\mathrm{mem}}(Y)
+
 r_{\mathrm{region}}(Y),
\end{equation}
where $r_{\mathrm{acc}}$ is the answer correctness reward. This reward formulation goes beyond merely rewarding the final answer's correctness by adding process-level rewards that supervise the use of visual evidence along the original reasoning trajectory.

\textbf{Group Relative Policy Optimization (GRPO).}
We optimize the policy $\pi_{\theta}$ with GRPO. For each training input, we sample a group of $G$ responses $\{Y_i\}_{i=1}^{G}$ from the old policy $\pi_{\theta_{\mathrm{old}}}$ and compute their rewards. The normalized advantage of response $Y_i$ is
\begin{equation}
\hat{A}_i = \frac{\mathcal{R}(Y_i,\mathcal{I},\mathcal{Q})-\mu_{\mathcal{R}}}{\sigma_{\mathcal{R}}},
\end{equation}
where $\mu_{\mathcal{R}}$ and $\sigma_{\mathcal{R}}$ are the mean and standard deviation of the rewards within the sampled group.

\begin{table*}[!t]
\vspace{-0.4em}
\centering
\captionsetup{skip=3pt}
\caption{Main results across seven benchmarks covering reasoning, general multimodal capability, and visual perception. Best results are shown in bold.}
\label{tab:main_results}
\renewcommand{\arraystretch}{0.93}
\resizebox{0.95\textwidth}{!}{
    \begin{tabular}{lccccccc}
    \toprule
    \multirow{2}{*}{\textbf{Models}} 
    & \multicolumn{3}{c}{\textbf{Mathematical \& Logical Reasoning}} 
    & \multicolumn{3}{c}{\textbf{Comprehensive Challenge}} 
    & \textbf{Perception} \\
    \cmidrule(lr){2-4} \cmidrule(lr){5-7} \cmidrule(lr){8-8}
     & MathVision & MathVista & LogicVista & MMVet & MMMB & MMStar & RealWorldQA \\
    \midrule

    \multicolumn{8}{l}{\textit{General MLLMs}} \\[-1pt]
    Qwen2.5-VL-3B~\citep{bai2025qwen2} & 20.90 & 51.90 & 40.49 & 52.98 & 80.60 & 54.73 & 65.22 \\
    Qwen2.5-VL-7B~\citep{bai2025qwen2} & 23.20 & 62.30 & 47.26 & 59.44 & 82.43 & 63.93 & 69.28 \\
    InternVL-4B~\citep{chen2024internvl} & 20.90 & 42.80 & 39.10 & 51.02 & 77.97 & 51.53 & 57.38 \\
    InternVL-8B~\citep{chen2024internvl} & 22.20 & 64.40 & 36.91 & 54.88 & 82.72 & 60.13 & 69.15 \\
    \midrule

    \multicolumn{8}{l}{\textit{Reasoning MLLMs}} \\[-1pt]
    Ocean-R1-3B~\citep{ming2025oceanr1} & 22.90 & 61.00 & 38.70 & 52.89 & 80.56 & 56.20 & 64.58 \\
    LMM-R1-3B~\citep{peng2025lmm} & 23.68 & 59.00 & 40.72 & 51.06 & 80.65 & 54.39 & 63.92 \\
    VLAA-Thinking-3B~\citep{chen2025sft} & 24.61 & 61.70 & 39.59 & 53.94 & 80.35 & 57.20 & 61.11 \\
    VLAA-Thinking-7B~\citep{chen2025sft} & 26.40 & 68.00 & 48.80 & 58.83 & \textbf{83.63} & 64.33 & 65.35 \\
    \midrule

    \multicolumn{8}{l}{\textit{Visual-Forgetting Mitigation MLLMs}} \\[-1pt]
    DeepSketcher-7B~\citep{zhang2025deepsketcher} & 22.03 & 69.10 & 45.19 & 69.54 & 78.68 & 57.33 & 65.88 \\
    TVC-7B~\citep{sun2025mitigating} & 22.70 & 62.10 & 38.71 & 60.41 & 74.29 & 55.73 & 62.21 \\
    \midrule

    \textbf{Remember-R1-3B (Ours)} & 24.68 & 65.50 & 42.95 & 63.44 & 80.95 & 59.54 & 65.88 \\
    \textbf{Remember-R1-7B (Ours)} & \textbf{26.64} & \textbf{69.80} & \textbf{49.05} & \textbf{72.37} & 83.38 & \textbf{64.73} & \textbf{69.67} \\
    \bottomrule
    \end{tabular}
}
\vspace{-0.4em}
\end{table*}

Let
\begin{equation}
\rho_i = \frac{\pi_{\theta}(Y_i \mid \mathcal{I},\mathcal{Q})}{\pi_{\theta_{\mathrm{old}}}(Y_i \mid \mathcal{I},\mathcal{Q})}
\end{equation}
be the probability ratio between the current and old policies. The clipped surrogate objective is
\begin{equation}
\mathcal{L}_i^{\mathrm{clip}}
=
\min\Big(
\rho_i \hat{A}_i,\;
\mathrm{clip}(\rho_i,1-\epsilon_{\mathrm{clip}},1+\epsilon_{\mathrm{clip}})\hat{A}_i
\Big).
\end{equation}
where $\epsilon_{\mathrm{clip}}$ is the clipping parameter.
The final optimization objective is
\begin{equation}
\mathcal{J}(\theta)
=
\mathbb{E}
\left[
\frac{1}{G}\sum_{i=1}^{G}\mathcal{L}_i^{\mathrm{clip}}
-
\beta D_{\mathrm{KL}}(\pi_{\theta}\parallel \pi_{\mathrm{ref}})
\right],
\end{equation}
where $\beta$ controls the strength of KL regularization and $\pi_{\mathrm{ref}}$ is the reference policy.

\section{Experiments}

\subsection{Experimental Settings}

\textbf{Models and Dataset.}
We evaluate \name\ at two model scales, using Qwen2.5-VL-3B and Qwen2.5-VL-7B~\citep{bai2025qwen2} as the base models. For training, we use ViRL39K~\citep{wang2025vl} and augment it with the visual keyword and key-region annotations described in Section~\ref{subsec:data_construction}.

\textbf{Benchmarks.}
We evaluate on seven benchmarks covering three categories: mathematical and logical reasoning, general multimodal capability, and visual perception. For mathematical and logical reasoning, we use MathVision~\citep{wang2024measuring}, MathVista~\citep{lu2024mathvista}, and LogicVista~\citep{xiao2024logicvista}, which require multi-step reasoning over visual inputs. For general multimodal capability, we use MMVet~\citep{yu2024mmvet}, MMMB~\citep{sun2025parrot}, and MMStar~\citep{chen2024we}, which cover a broad range of multimodal tasks beyond reasoning-focused settings. For visual perception, we use RealWorldQA~\citep{xai2024grok15v}, which emphasizes recognition of fine-grained visual details in realistic scenarios. We follow the standard evaluation protocols of the corresponding benchmarks.

\textbf{Compared Methods.}
We compare \name\ with three groups of open-source baselines: (1) general MLLMs, including Qwen2.5-VL-3B/7B~\citep{bai2025qwen2} and InternVL-4B/8B~\citep{chen2024internvl}; (2) reasoning-oriented MLLMs, including Ocean-R1-3B~\citep{ming2025oceanr1}, LMM-R1-3B~\citep{peng2025lmm}, and VLAA-Thinking-3B/7B~\citep{chen2025sft}; and (3) prior Visual-Forgetting Mitigation MLLMs, including DeepSketcher-7B~\citep{zhang2025deepsketcher} and TVC-7B~\citep{sun2025mitigating}.

\textbf{Implementation Details.}
All training experiments are conducted on 8 NVIDIA L20 GPUs. We set the learning rate to $1\times10^{-5}$ and the GRPO group size to 8. We use the same hyperparameter settings for both model scales to test whether the proposed method remains effective across scales.

\subsection{Main Results}

\textbf{Overview.}
Table~\ref{tab:main_results} shows that \name\ consistently improves over the corresponding Qwen2.5-VL base models at both the 3B and 7B scales. Taken together, these results support the overall reward design: supervising matched annotated visual-keyword coverage, later-step visual persistence, and question-relevant attention on the original reasoning trajectory leads to stronger performance across reasoning, general multimodal evaluation, and visual perception. The gains are especially clear on MathVista, MMVet, and MMStar, while the remaining benchmarks are consistently maintained or improved.

\begin{table*}[!t]
\centering
\caption{Ablation study on seven benchmarks. We compare the base model, an accuracy-only GRPO variant trained with the correctness reward $r_{\mathrm{acc}}$, and variants of \name\ with one process-level reward removed at a time. Removing any reward leads to performance degradation, while the full model achieves the strongest and most consistent results across benchmarks.}
\label{tab:ablation_results}
\resizebox{0.95\textwidth}{!}{
    \begin{tabular}{lccccccc}
    \toprule
    \textbf{Models} & \textbf{MathVision} & \textbf{MathVista} & \textbf{LogicVista} & \textbf{MMVet} & \textbf{MMMB} & \textbf{MMStar} & \textbf{RealWorldQA} \\
    \midrule
    Qwen2.5-VL-3B~\citep{bai2025qwen2} & 20.90 & 51.90 & 40.49 & 52.98 & 80.60 & 54.73 & 65.22 \\
    Qwen2.5-VL-3B-GRPO ($r_{\mathrm{acc}}$) & 21.05 & 60.20 & 40.79 & 59.98 & 80.70 & 56.20 & 64.75 \\
    \midrule
    Remember-R1 w/o $r_{\mathrm{voc}}$ & 21.71 & 62.00 & 41.62 & 61.88 & 80.80 & 57.86 & 65.35 \\
    Remember-R1 w/o $r_{\mathrm{mem}}$ & 24.27 & 62.20 & 41.62 & 63.36 & 80.75 & 57.86 & 65.31 \\
    Remember-R1 w/o $r_{\mathrm{region}}$ & 21.05 & 61.70 & 40.82 & 62.50 & 80.70 & 58.20 & 63.79 \\
    \midrule
    \textbf{Remember-R1-3B} & \textbf{24.68} & \textbf{65.50} & \textbf{42.95} & \textbf{63.44} & \textbf{80.95} & \textbf{59.54} & \textbf{65.88} \\
    \bottomrule
    \end{tabular}
}
\end{table*}

\textbf{Results on reasoning-intensive benchmarks.}
On MathVision, MathVista, and LogicVista, \name\ improves over the corresponding Qwen2.5-VL base models at both scales. Specifically, the 3B/7B gains are $+3.78/+3.44$ on MathVision, $+13.60/+7.50$ on MathVista, and $+2.46/+1.79$ on LogicVista. Compared with recent methods designed to mitigate visual forgetting, \name\ remains competitive overall and achieves stronger results on several benchmarks. These results show that the benefits extend to reasoning-intensive tasks while also appearing on broader multimodal evaluations.

\textbf{Results on general multimodal and perception benchmarks.}
Beyond reasoning-focused settings, \name\ performs competitively on general multimodal and perception benchmarks. Relative to the corresponding Qwen2.5-VL base models, it improves MMVet and MMStar at both scales, suggesting that the benefits of process-level supervision extend beyond reasoning-specific evaluations. It also compares favorably with recent visual-forgetting mitigation methods and reasoning MLLMs on several broader benchmarks. On RealWorldQA, accuracy improves slightly at both model scales, indicating that the reasoning gains do not come at the expense of fine-grained visual perception. Overall, these comparisons suggest that direct supervision on the original reasoning trajectory can improve reasoning-intensive behavior while preserving general multimodal capability and perception quality.

\subsection{Visual Grounding Analysis}

\begin{figure}[t]
    \centering
    \includegraphics[width=1\linewidth]{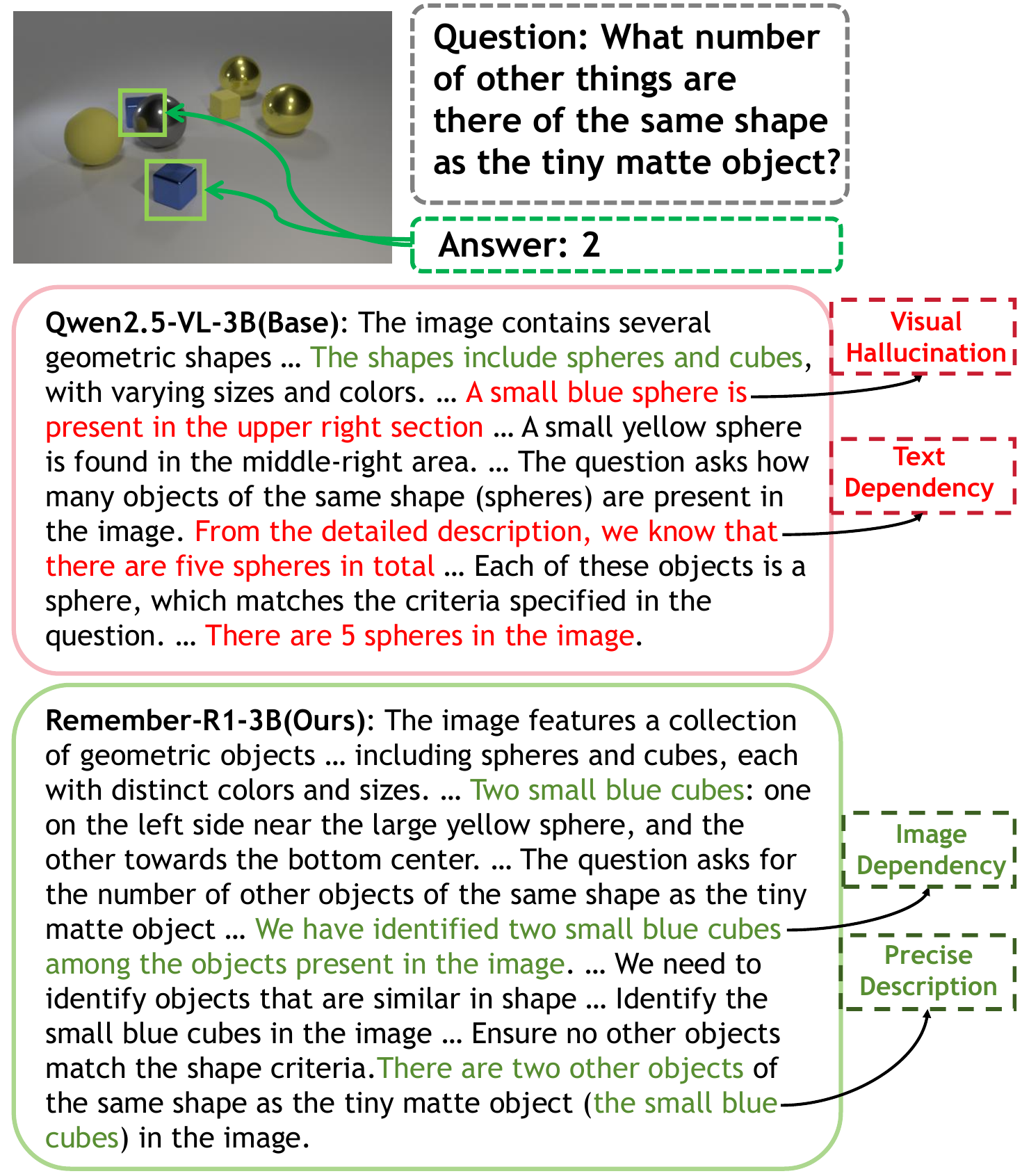}
    \vspace{-1.5em}
    \caption{Qualitative comparison on geometric reasoning. The base model drifts to an incorrect text-based count, whereas \name\ uses shape, color, and spatial evidence to answer correctly.}
    \label{fig:quality}
\end{figure}

\textbf{Qualitative Analysis.}
Figure~\ref{fig:quality} illustrates how \name\ remains grounded in visual evidence during geometric reasoning. In this example, the base model hallucinates a nonexistent blue sphere and then relies on this unsupported description to count five spheres, producing an incorrect answer. By contrast, \name\ explicitly tracks the relevant shape, color, and spatial cues, identifies the two cube-shaped objects that match the query, and returns the correct answer. This comparison illustrates how \name, by sustaining use of question-relevant visual evidence, can prevent a reasoning trajectory from drifting toward unsupported textual assumptions.

\begin{figure}[t]
    \centering
    \includegraphics[width=0.80\linewidth]{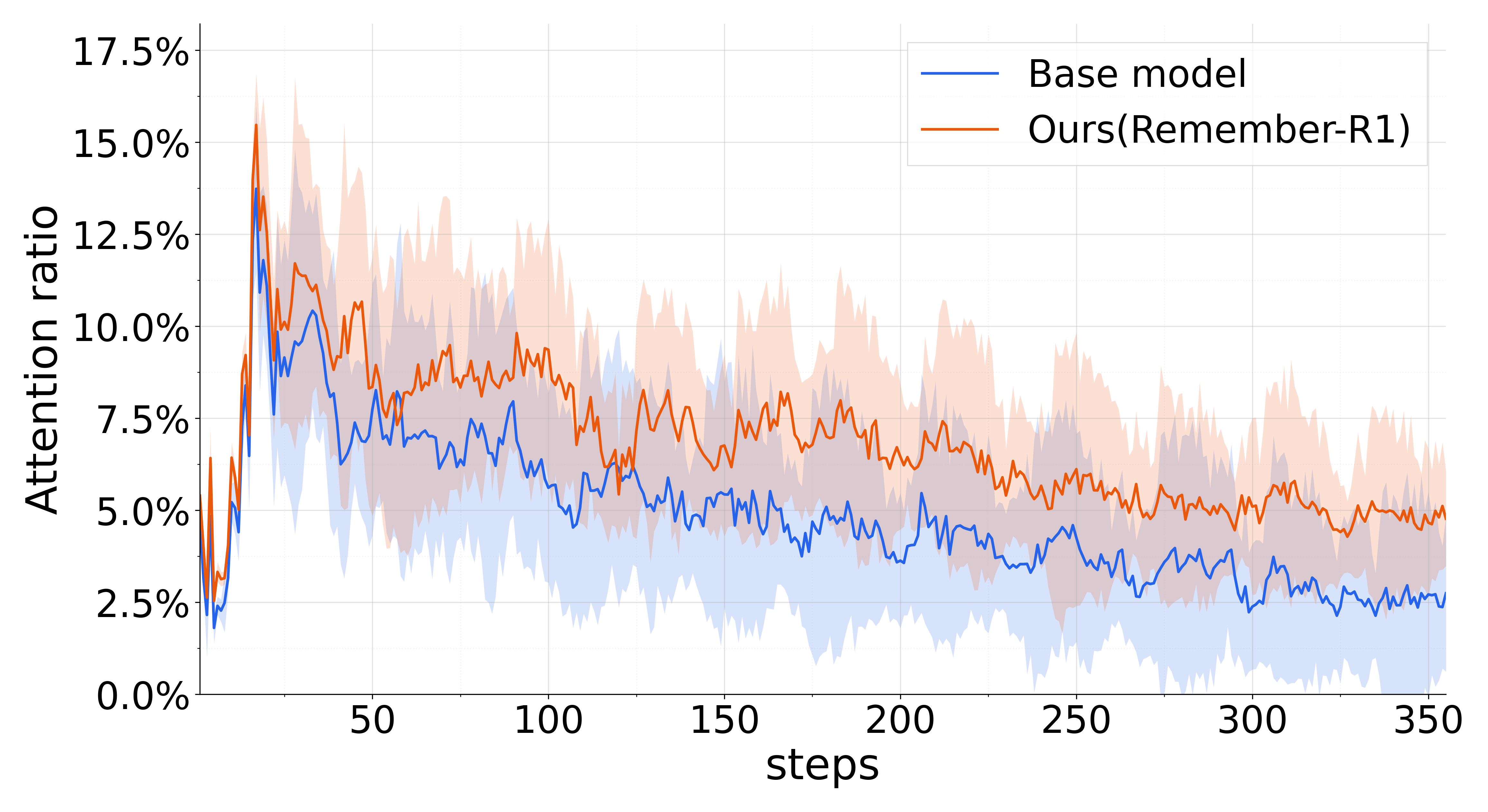}
    \centerline{\small (a) Attention ratio on LogicVista.}

    \includegraphics[width=0.80\linewidth]{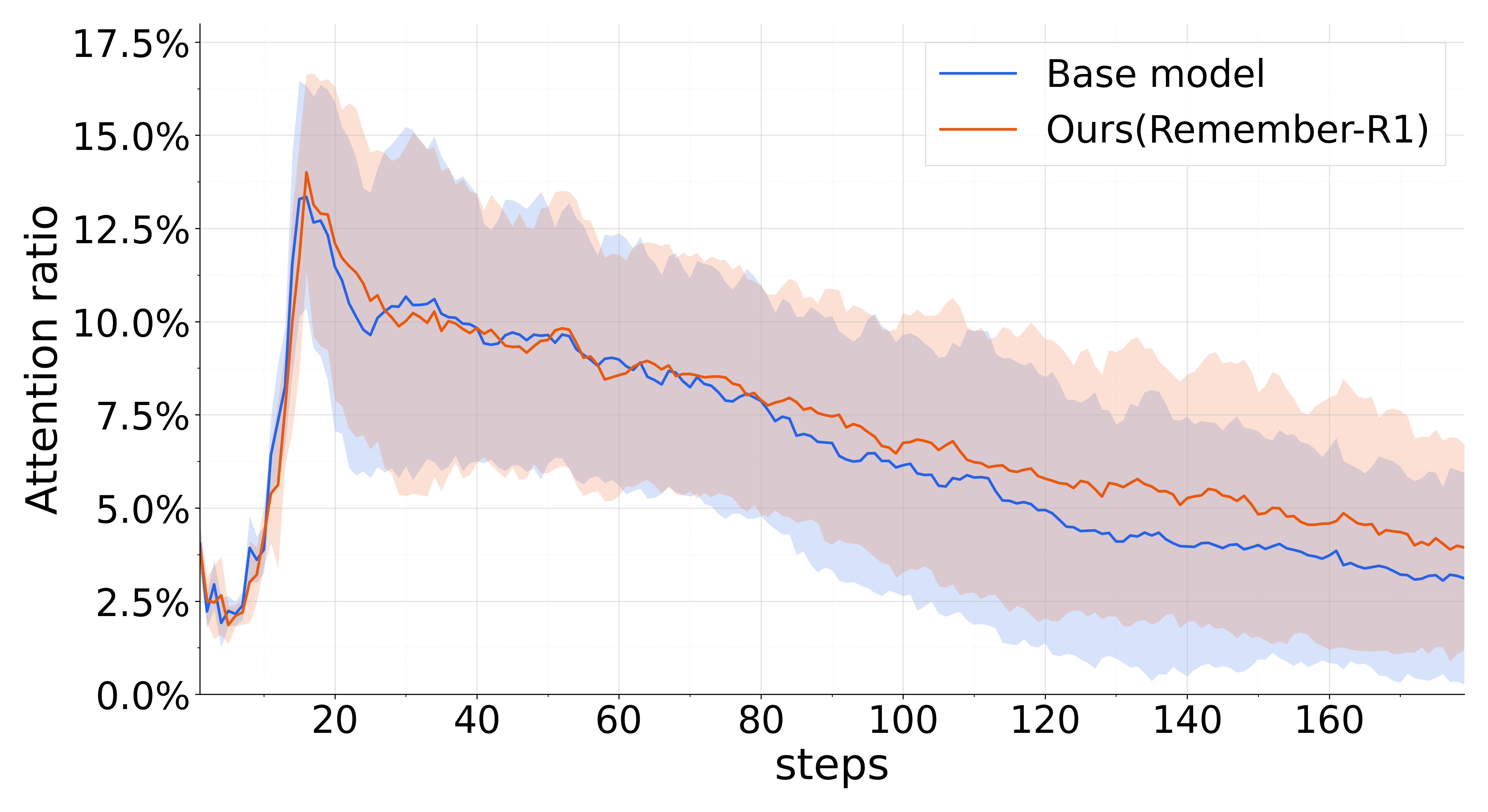}
    \centerline{\small (b) Attention ratio on MMStar.}
    \caption{Mean visual attention ratio across reasoning steps for the base model and \name\ on LogicVista~\citep{xiao2024logicvista} and MMStar~\citep{chen2024we}.}
    \label{fig:atten_ratio}
\end{figure}

\textbf{Attention Ratio Analysis.}
Figure~\ref{fig:atten_ratio} compares the mean attention ratio assigned to visual tokens across reasoning steps on LogicVista and MMStar. On both benchmarks, the attention ratio decreases over time for both models, indicating that long-form generation naturally weakens reliance on visual evidence. Compared with the base model, however, \name\ shows a consistently slower decline, with the gap becoming more pronounced in the middle and later stages. This pattern suggests that \name\ more effectively preserves reliance on visual evidence as the reasoning trajectory unfolds, rather than shifting too quickly toward previously generated text. That this same trend holds on both a reasoning and a general multimodal benchmark indicates that the behavior generalizes across distinct evaluation settings.

\subsection{Ablation Study}
\begin{figure}[t]
    \centering
    \includegraphics[width=0.44\linewidth]{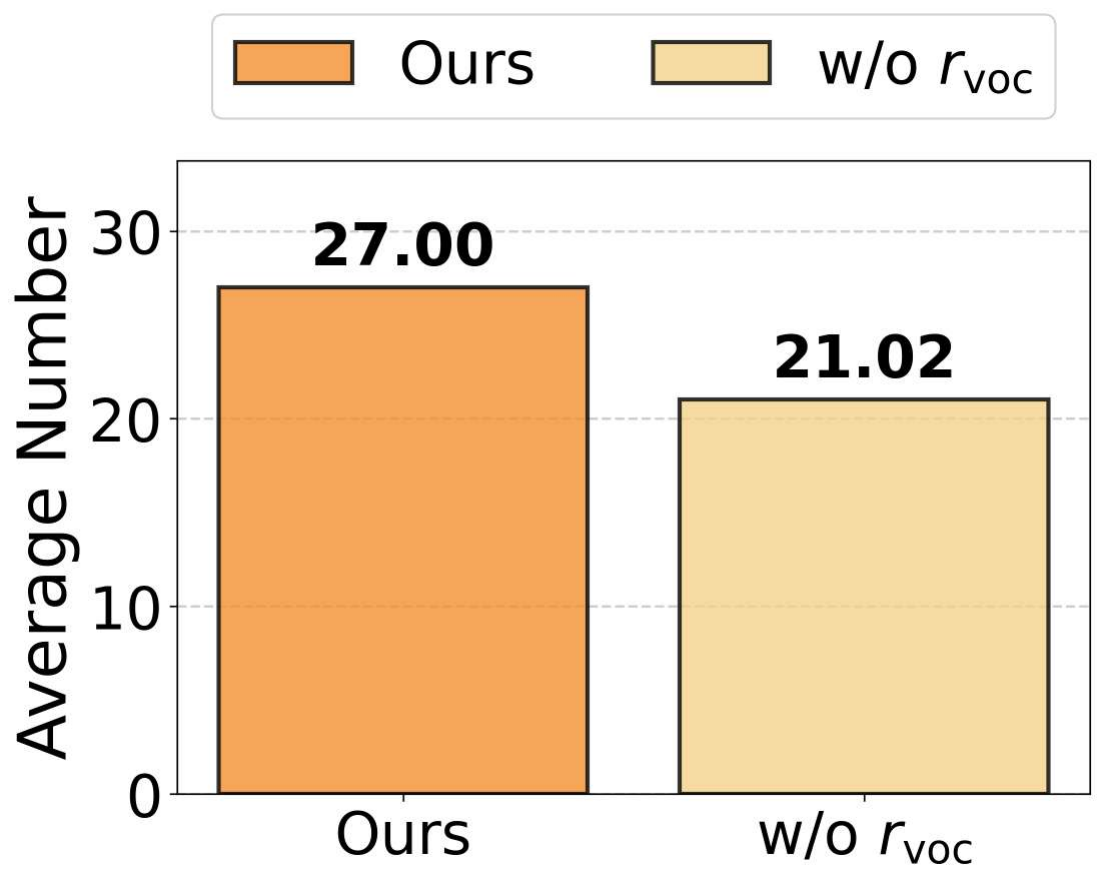}
    \hfill
    \includegraphics[width=0.49\linewidth]{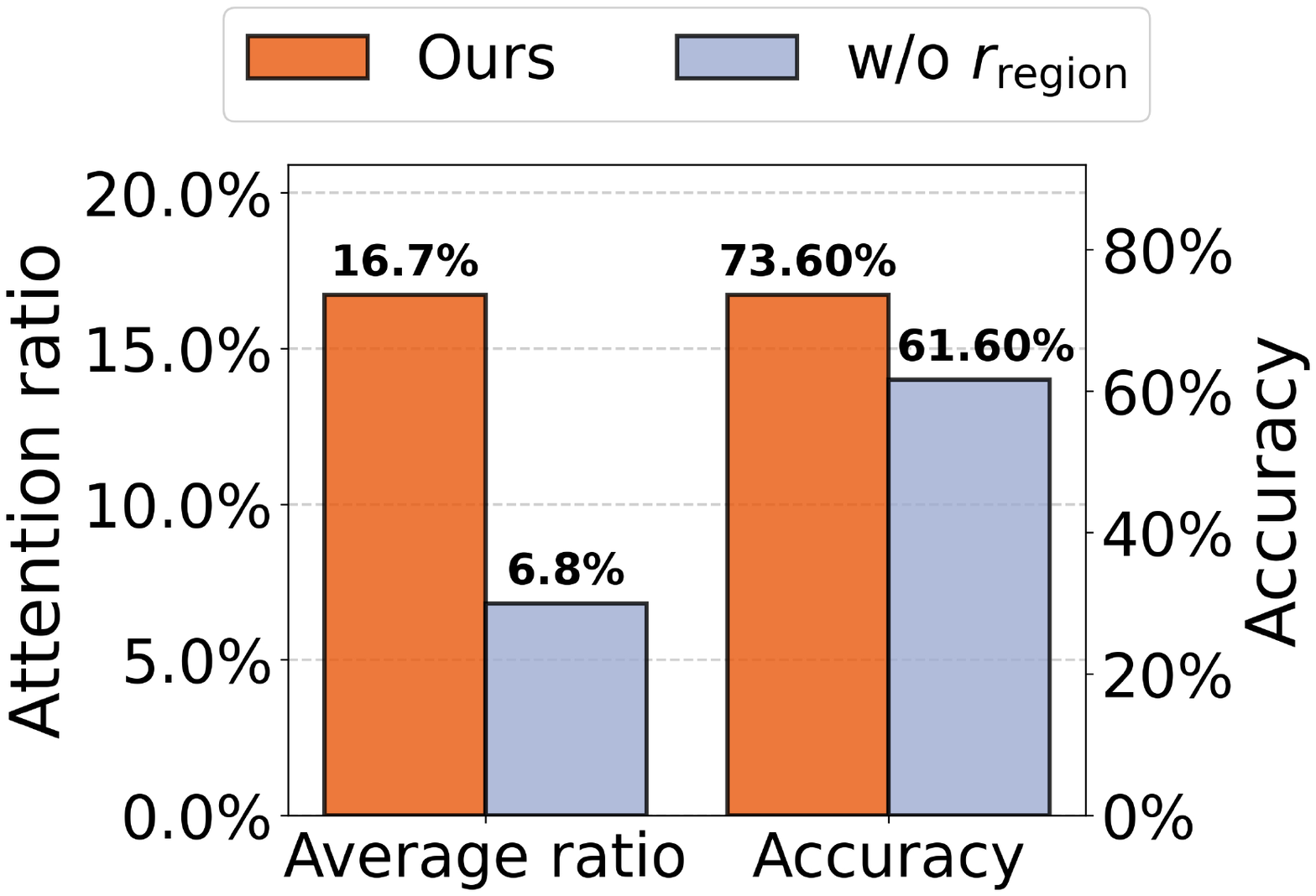}
    \par
    \makebox[0.44\linewidth][c]{\small (a) w/o $r_{\mathrm{voc}}$}
    \hfill
    \makebox[0.49\linewidth][c]{\small (c) w/o $r_{\mathrm{region}}$}

    \vspace{0.2em}

    \includegraphics[width=\linewidth]{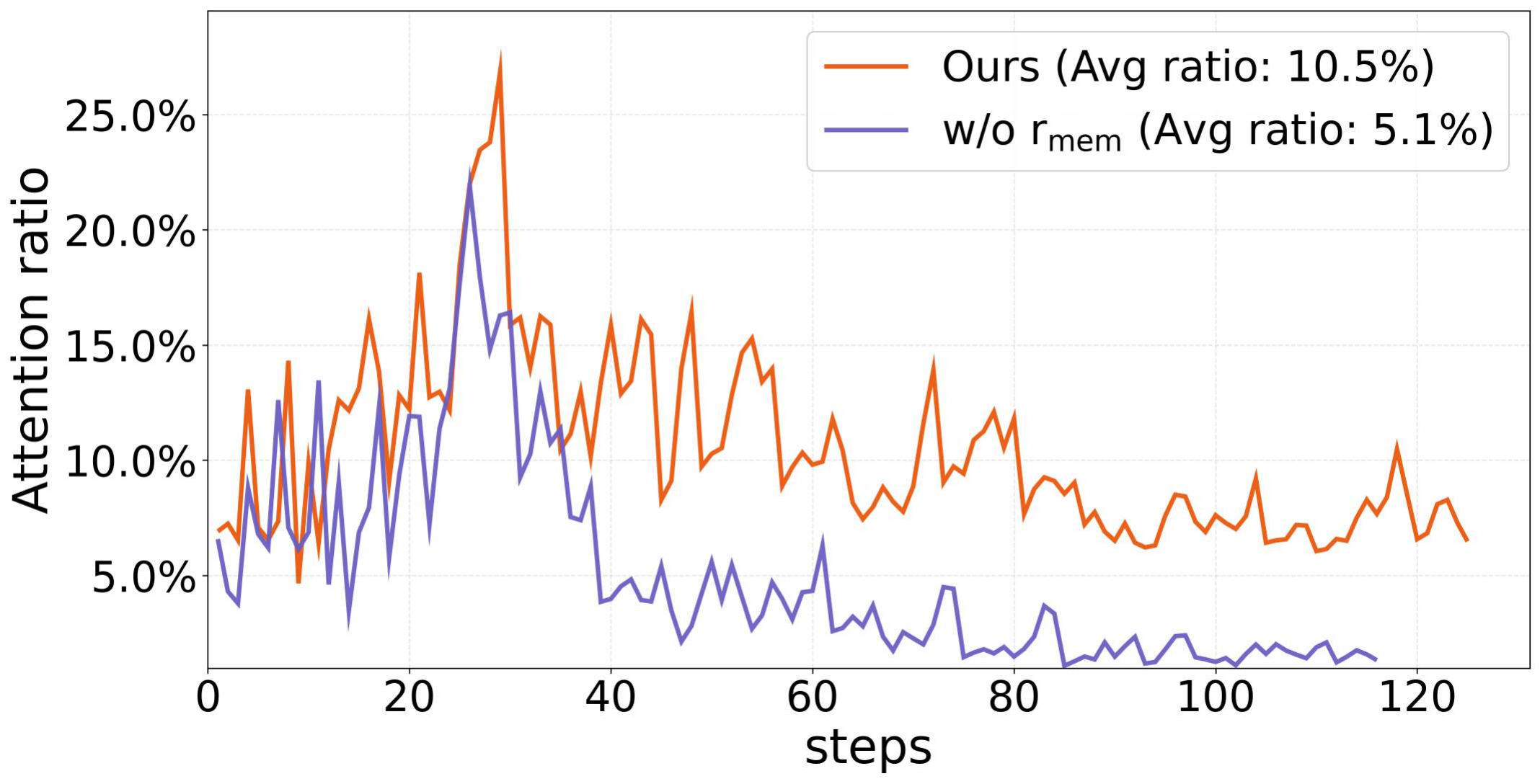}
    \centerline{\small (b) w/o $r_{\mathrm{mem}}$}

    \caption{Effects of removing the three process-level rewards: (a) the matched annotated visual-keyword count decreases without $r_{\mathrm{voc}}$; (b) visual attention decays faster without $r_{\mathrm{mem}}$; and (c) key-region attention decreases without $r_{\mathrm{region}}$.}
    \label{fig:ablation_studies}
    \vspace{-1em}
\end{figure}

\textbf{Overall ablation results.}
Table~\ref{tab:ablation_results} and Figure~\ref{fig:ablation_studies} evaluate the three-part reward design. Removing any reward degrades performance on multiple benchmarks, whereas the full model performs best overall. These results confirm that matched annotated visual-keyword coverage, later-step visual persistence, and question-relevant attention provide complementary benefits.

\textbf{Effectiveness of $r_{\mathrm{voc}}$.}
Removing $r_{\mathrm{voc}}$ reduces the number of matched annotated visual keywords in the response, as shown by the lower visual-keyword count in Figure~\ref{fig:ablation_studies}(a). This change is accompanied by weaker results on multiple benchmarks in Table~\ref{tab:ablation_results}, especially on reasoning-heavy settings such as MathVista. Overall, degradation is most evident on reasoning-oriented benchmarks, while the remaining evaluations show smaller, consistent changes. These results suggest that encouraging coverage of visual keywords is an important part of the overall improvement. Without this reward, the model tends to produce more abstract or text-dominated reasoning, which can drift from the underlying image facts.

\textbf{Effectiveness of $r_{\mathrm{mem}}$.}
Removing $r_{\mathrm{mem}}$ leads to a noticeably faster decline in visual attention over generation steps in Figure~\ref{fig:ablation_studies}(b), while Table~\ref{tab:ablation_results} shows corresponding performance drops across several benchmarks. Across benchmarks, the effect varies in magnitude, but the overall trend favors the complete reward configuration. This result indicates that preserving reliance on visual evidence into later reasoning steps is important for maintaining performance, especially when the reasoning chain is long. In its absence, the model shifts dependence toward previously generated text earlier in the generation process, weakening the visual grounding needed for subsequent reasoning.

\textbf{Effectiveness of $r_{\mathrm{region}}$.}
Removing $r_{\mathrm{region}}$ substantially reduces attention aligned with question-relevant regions in Figure~\ref{fig:ablation_studies}(c), together with clear accuracy degradation. Consistently, Table~\ref{tab:ablation_results} shows broader performance drops after removing this reward. The performance decrease appears across reasoning-intensive, general multimodal, and perception benchmarks, showing that the effect is not task-specific. This suggests that maintaining overall visual attention alone is insufficient; without this reward, the model may distribute attention to less informative areas, and $r_{\mathrm{region}}$ helps maintain focused and consistent use of visual cues throughout the reasoning trajectory.

\textbf{Comparison with accuracy-only RL.}
Compared with the GRPO variant trained only with the correctness reward $r_{\mathrm{acc}}$, the full model performs better on most benchmarks in Table~\ref{tab:ablation_results}. This comparison indicates that answer-level supervision alone is insufficient. Process-level rewards provide additional supervision on matched annotated visual-keyword coverage and on how visual attention is maintained and localized along the reasoning trajectory. While accuracy-only RL can improve final answers to some extent, it does not explicitly shape the underlying reasoning process, which is crucial for tasks that demand sustained visual grounding.

\section{Conclusion}
In this paper, we introduced \name, a reinforcement learning framework that mitigates visual forgetting in long-context multimodal reasoning through process-level supervision without altering inference. Its visual vocabulary, memory, and key-region rewards jointly improve visual grounding throughout the reasoning trajectory. Experiments across reasoning, multimodal understanding, and perception benchmarks show consistent gains across model scales, highlighting process-level reinforcement learning as a promising approach to long-horizon multimodal reasoning. The attention and ablation analyses further indicate that these gains arise from sustaining and localizing task-relevant visual evidence, rather than relying solely on final-answer supervision.

\begin{acks}
This work was supported by the National Natural Science Foundation of China (Grant Nos. 62472359 and 62372379), Xi'an's Key Industrial Chain Core Technology Breakthrough Project: AI Core Technology Breakthrough (Grant No. 24ZDCYJSGG0003), the National Key Research and Development Program of China (Grant No. 2023YFB2703700), and the Research Grants Council of HKSAR under grant number AoE/E-601/24-N.
\end{acks}

\bibliographystyle{acm_reference_format}
\bibliography{references}

\end{document}